\documentclass[conference]{IEEEtran}
\IEEEoverridecommandlockouts
\usepackage{cite}
\usepackage{amsmath,amssymb,amsfonts}
\usepackage{algorithmic}
\usepackage{graphicx}
\usepackage{textcomp}
\usepackage{xcolor}
\def\BibTeX{{\rm B\kern-.05em{\sc i\kern-.025em b}\kern-.08em
    T\kern-.1667em\lower.7ex\hbox{E}\kern-.125emX}}
    
\newtheorem{definition}{Definition}

\usepackage{tikz}
\usetikzlibrary{positioning, shapes.geometric, arrows.meta}
\usepackage{wrapfig}
\usepackage{subfigure}

\newcommand\w{\omega}
\newcommand{\vars}{\texttt{vars}}
\newcommand{\iuq}{\overline{\forall}\,}
\newcommand{\lit}{\ell}
\newcommand{\inst}{{\cal I}}
\newcommand\NOT[1]{\overline{#1}}
\newcommand\AND{\cdot}
\newcommand\OR{+}
\newcommand\BAND{\prod}
\newcommand\BOR{\sum}

\newcommand{\Qa}{\text{Q1}}
\newcommand{\Qb}{\text{Q2}}
\newcommand{\Qc}{\text{Q3}}

\newcommand\shrink[1]{}
\newcommand{\eql}[2]{{#1}\mathord=\mathord{#2}}
\newcommand{\IN}{\,\mathord\in\mathord}
\newcommand\yes{{\footnotesize\textsc{yes}}}
\newcommand\no{{\footnotesize\textsc{no}}}
\newcommand\Pos{{\footnotesize \textsc{+ve}}}
\newcommand\Neg{{\footnotesize \textsc{--ve}}}
\newcommand\Age{{\footnotesize \textsc{Age}}}
\newcommand\BMI{{\footnotesize \textsc{BMI}}}
\newcommand\Weight{{\footnotesize \textsc{Weight}}}
\newcommand\BType{{\footnotesize \textsc{Btype}}}
\newcommand\Db{{\footnotesize \textsc{Diabetes}}}
\newcommand\yy{{\footnotesize \textsc{y}}}
\newcommand\nn{{\footnotesize \textsc{n}}}
\newcommand\UWeight{{\footnotesize \textsc{under}}}
\newcommand\OWeight{{\footnotesize \textsc{over}}}
\newcommand\Nom{{\footnotesize \textsc{norm}}}
\newcommand\tA{{\footnotesize \textsc{A}}}
\newcommand\tB{{\footnotesize \textsc{B}}}
\newcommand\tAB{{\footnotesize \textsc{AB}}}
\newcommand\tO{{\footnotesize \textsc{O}}}

\begin{document}

\title{Logic for Explainable AI  \\
\thanks{This work has been partially supported by NSF grant \#ISS-1910317.}
}

\author{\IEEEauthorblockN{Adnan Darwiche}
\IEEEauthorblockA{\textit{Computer Science Department} \\
\textit{University of California}\\
Los Angeles, USA \\
darwiche@cs.ucla.edu}
}

\IEEEoverridecommandlockouts
\IEEEpubid{\makebox[\columnwidth]{\footnotesize In $38^{\text {th}}$ Annual ACM/IEEE Symposium on Logic in Computer Science, 2023\hfill} \hspace{\columnsep}\makebox[\columnwidth]{ }}
   
\maketitle

\begin{abstract}
A central quest in explainable AI relates to understanding the decisions made
by (learned) classifiers. There are three dimensions of this understanding that have been receiving 
significant attention in recent years. 
The first dimension
relates to characterizing conditions on instances that are necessary and sufficient for decisions,
therefore providing abstractions of instances that can be viewed as the ``reasons behind decisions.''
The next dimension
relates to characterizing minimal conditions that are sufficient for a decision,
therefore identifying maximal aspects of the instance that are irrelevant to the decision.
The last dimension 
relates to characterizing minimal conditions that are necessary for a decision, 
therefore identifying minimal perturbations to the instance that yield alternate decisions. 
We discuss in this tutorial a comprehensive, semantical and computational theory of explainability 
along these dimensions which is based on some recent developments in symbolic logic.
The tutorial will also discuss how this theory is particularly applicable to
non-symbolic classifiers such as those based on Bayesian networks, decision trees,
random forests and some types of neural networks.
\end{abstract}


\section{Introduction}

\begin{figure}[tb]
\centering
\subfigure[Na\"ive Bayes classifier]{\label{fig:nbc}\includegraphics[width=5.6cm]{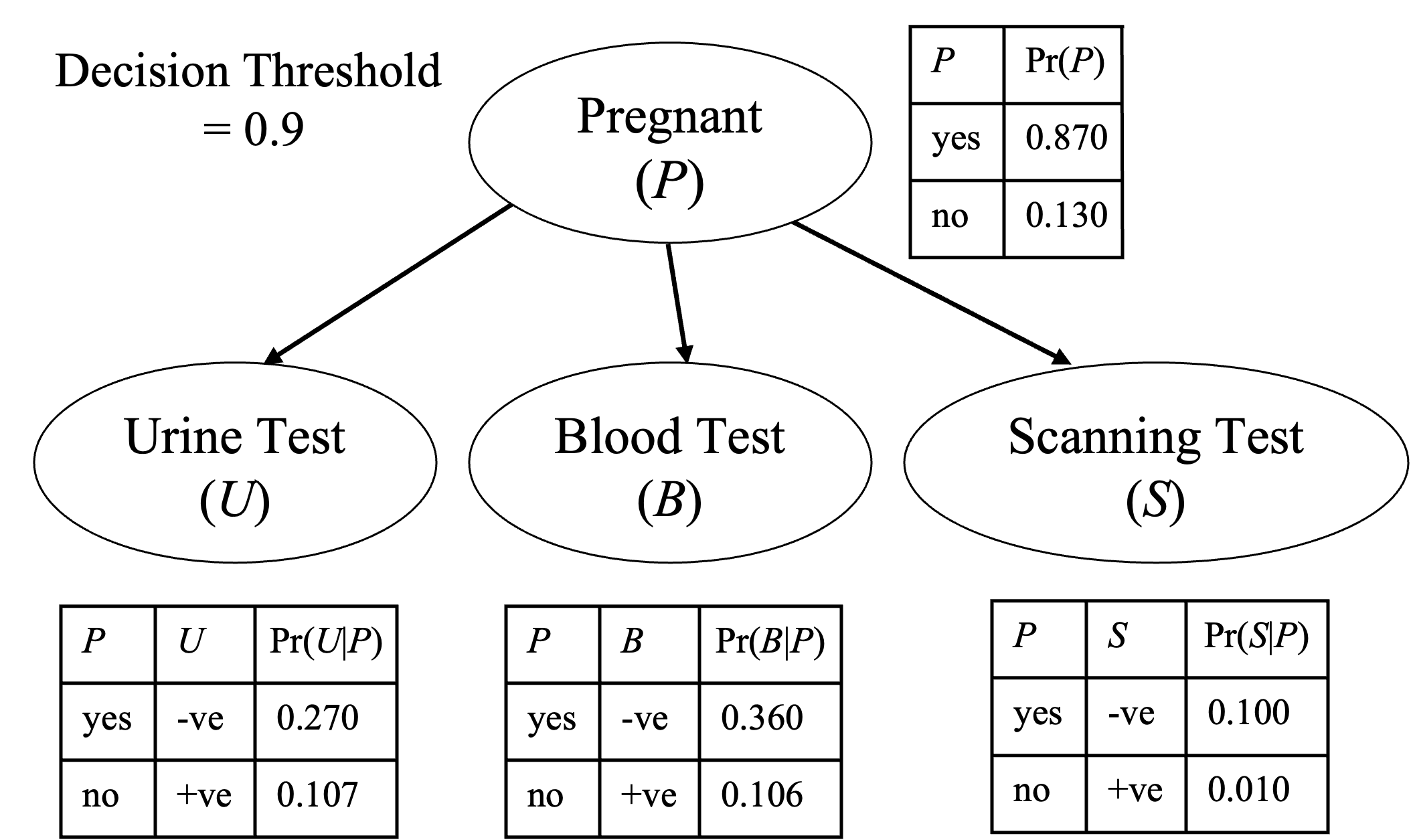}}\quad
\subfigure[decision graph]{\label{fig:nbc-dgraph}\includegraphics[width=2.9cm]{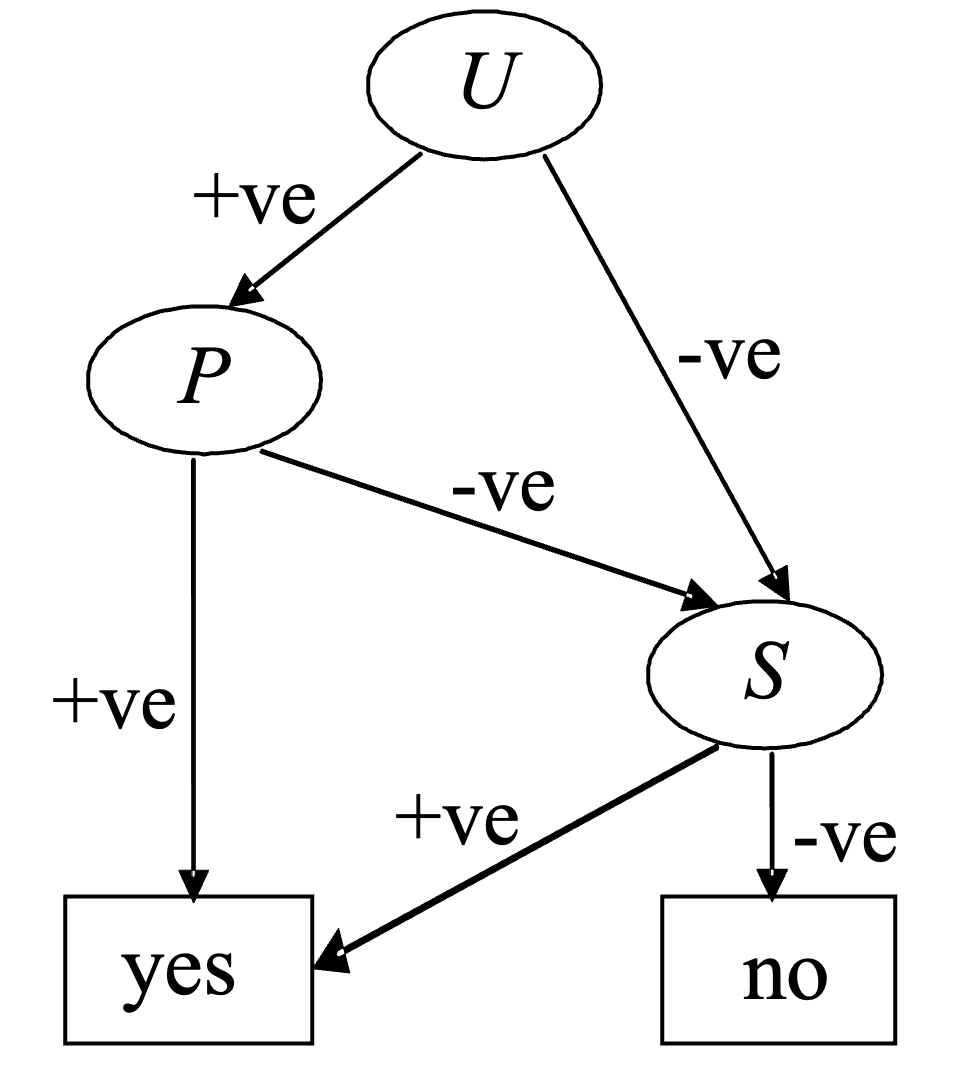}}
\caption{A Na\"ive Bayes classifier, from~\cite{DBLP:conf/uai/ChanD03}, with binary features $U, B, S$ and classes $\eql{P}{\yes}$, $\eql{P}{\no}$.
The decision graph represents the same classifier as it will make the exact same decision on every instance.}
\label{fig:nbc2dgraph}
\end{figure}

\begin{figure}[tb]
\centering
\subfigure[decision tree]{\label{fig:dtree-c} \resizebox{0.1585\textwidth}{!}{%
        \begin{tikzpicture}[
        roundnode/.style={text width = 0.55cm, circle ,draw=black, thick, text badly centered},
        squarednode/.style={rectangle, draw=black, thick, text badly centered},
        ]
        \node[squarednode]     (AGE1)                              {\large \Age};
        \node[squarednode]     (BMI2)       [below=of AGE1, xshift = -0.8cm, yshift = 0.4cm] {\large \BMI};
        \node[squarednode]     (AGE2)       [below=of AGE1, xshift = 0.8cm, yshift = -1.2cm] {\large \Age};
        \node[roundnode]       (NO3)   [, below=of BMI2, xshift = -0.5cm, yshift = 0.4cm] {\large \no};
        \node[roundnode]       (YES3)   [below=of BMI2, xshift = 0.5cm, yshift = 0.4cm] {\large \yes};
        \node[squarednode]     (BMI31)       [below=of AGE2, xshift = -1.45cm, yshift = 0.4cm] {\large \BMI};
        \node[squarednode]     (BMI32)       [below=of AGE2, xshift = 0.15cm, yshift = -1.2cm] {\large \BMI};
        \node[roundnode]       (NO41)   [below=of BMI31, xshift = -0.5cm, yshift = 0.4cm] {\large \no};
        \node[roundnode]       (YES42)   [below=of BMI31, xshift = 0.5cm, yshift = 0.4cm] {\large \yes};
        \node[roundnode]       (NO43)   [below=of BMI32, xshift = -0.1cm, yshift = 0.4cm] {\large \no};
        \node[roundnode]       (YES44)   [below=of BMI32, xshift = -1.2cm, yshift = 0.4cm] {\large \yes};
        
        \draw[-latex, thick] (AGE1.240) -- node [anchor = center, xshift = -6mm, yshift = 1mm] {$<18$} (BMI2.north);
        \draw[-latex, thick] (AGE1.300) -- node [anchor = center, xshift = 6mm, yshift = 1mm] {$\ge18$} (AGE2.north);
        \draw[-latex, thick] (BMI2.240) -- node [anchor = center, xshift = -6mm] {$<30$} (NO3.north);
        \draw[-latex, thick] (BMI2.300) -- node [anchor = center, xshift = 6mm] {$\ge30$} (YES3.north);
        \draw[-latex, thick] (AGE2.240) -- node [anchor = center, xshift = -5mm, yshift = 1mm] {$<40$} (BMI31.north);
        \draw[-latex, thick] (AGE2.300) -- node [anchor = center, xshift = 5mm, yshift = 1mm] {$\ge40$} (BMI32.north);
        \draw[-latex, thick] (BMI31.240) -- node [anchor = center, xshift = -5mm] {$<27$} (NO41.north);
        \draw[-latex, thick] (BMI31.300) -- node [anchor = center, xshift = 5mm] {$\ge27$} (YES42.north);
        \draw[-latex, thick] (BMI32.240) -- node [anchor = center, xshift = -6mm] {$\ge25$} (YES44.north);
        \draw[-latex, thick] (BMI32.300) -- node [anchor = center, xshift = 5mm] {$<25$} (NO43.north);
        \end{tikzpicture}
}%
}\quad
\subfigure[discretized decision tree]{\label{fig:dtree-d} \resizebox{0.20\textwidth}{!}{%
        \begin{tikzpicture}[
        roundnode/.style={text width = 0.55cm, circle ,draw=black, thick, text badly centered},
        squarednode/.style={rectangle, draw=black, thick, text badly centered},
        ]
        \node[squarednode]     (AGE1)                              {\large \Age};
        \node[squarednode]     (BMI2)       [below=of AGE1, xshift = -1.05cm, yshift = 0.4cm] {\large \BMI};
        \node[squarednode]     (AGE2)       [below=of AGE1, xshift = 0.8cm, yshift = -1.2cm] {\large \Age};
        \node[roundnode]       (NO3)   [, below=of BMI2, xshift = -0.5cm, yshift = 0.4cm] {\large \no};
        \node[roundnode]       (YES3)   [below=of BMI2, xshift = 0.5cm, yshift = 0.4cm] {\large \yes};
        \node[squarednode]     (BMI31)       [below=of AGE2, xshift = -2.85cm, yshift = 0.4cm] {\large \BMI};
        \node[squarednode]     (BMI32)       [below=of AGE2, xshift = 0.2cm, yshift = -1.2cm] {\large \BMI};
        \node[roundnode]       (NO41)   [below=of BMI31, xshift = -0.5cm, yshift = 0.4cm] {\large \no};
        \node[roundnode]       (YES42)   [below=of BMI31, xshift = 0.5cm, yshift = 0.4cm] {\large \yes};
        \node[roundnode]       (YES44)   [below=of BMI32, xshift = -1.3cm, yshift = 0.4cm] {\large \yes};
        \node[roundnode]       (NO43)   [below=of BMI32, xshift = 0cm, yshift = 0.4cm] {\large \no};
        
        \draw[-latex, thick] (AGE1.240) -- node [anchor = center, xshift = -3mm, yshift = 1mm] {$a_1$} (BMI2.north);
        \draw[-latex, thick] (AGE1.300) -- node [anchor = center, xshift = 3mm, yshift = 8mm] {$a_2, a_3$} (AGE2.north);
        \draw[-latex, thick] (BMI2.240) -- node [anchor = center, xshift = -9mm] {$b_1, b_2, b_3$} (NO3.north);
        \draw[-latex, thick] (BMI2.300) -- node [anchor = center, xshift = 4mm] {$b_4$} (YES3.north);
        \draw[-latex, thick] (AGE2.240) -- node [anchor = center, xshift = 3mm, yshift = -2mm] {$a_2$} (BMI31.north);
        \draw[-latex, thick] (AGE2.300) -- node [anchor = center, xshift = 3mm, yshift = 1mm] {$a_3$} (BMI32.north);
        \draw[-latex, thick] (BMI31.240) -- node [anchor = center, xshift = -6mm] {$b_1, b_2$} (NO41.north);
        \draw[-latex, thick] (BMI31.300) -- node [anchor = center, xshift = 6mm] {$b_3, b_4$} (YES42.north);
        \draw[-latex, thick] (BMI32.240) -- node [anchor = center, xshift = -9mm,yshift = 1mm] {$b_2, b_3, b_4$} (YES44.north);
        \draw[-latex, thick] (BMI32.300) -- node [anchor = center, xshift = 3mm] {$b_1$} (NO43.north);
        \end{tikzpicture}
}%
}
\caption{A decision tree, from~\cite{corr/abs-2304.14760}, with numeric features $\Age$, $\BMI$ and classes $\yes$, $\no$.
$\Age$ is discretized into three intervals $[0, 18), [18, 40)$ and $[40, \infty)$ so it can be treated as a discrete variable with respective values \(a_1, a_2, a_3\). 
$\BMI$ is discretized into four intervals $[0, 25), [25, 27), [27, 30), [30, \infty)$ so it can be treated as a discrete variable with respective values $b_1, b_2, b_3, b_4$.
}
\label{fig:decision-tree}
\end{figure}
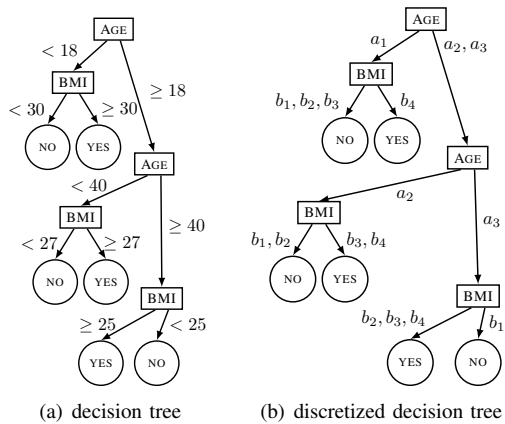

Explaining the decisions of (learned) classifiers is perhaps the most studied task in the area of explainable AI.
A classifier can take different forms---like a decision tree, random forest, Bayesian network
or a neural network---but it is essentially a function that maps \textit{instances} to a finite number of \textit{classes.}
When a classifier maps an instance to a class, we say it has made a \textit{decision} on that instance.
Each classifier has a set of \textit{features} which are discrete or numeric variables. An instance is generated by
assigning a value to each feature. 
Consider the Na\"ive Bayes classifier in Fig.~\ref{fig:nbc} which has three binary features 
($U$, $B$, $S$), representing medical tests, which generate eight instances. This classifier has
two classes ($\eql{P}{\yes}$, $\eql{P}{\no}$) corresponding to whether the patient is pregnant or not. 
Given an instance which represents the test results of a patient, this classifier first computes
the distribution on variable $P$ given these results and then assigns the class $\eql{P}{\yes}$ to the instance iff this probability is no
less than \(0.90\) (called the classification threshold). This classifier could have been learned from
data and it makes decisions by performing probabilistic reasoning, but it is in essence a function
that maps a finite number of instances to classes. Fig.~\ref{fig:dtree-c} depicts another classifier 
in the form of a decision tree. 
This one has two numeric features ($\Age$, $\BMI$) and, hence, an infinite number of instances which are mapped by
the classifier into one of two classes ($\yes$, $\no$).

The goal of this tutorial is to discuss a theory based on symbolic logic for explaining the decisions of such classifiers.
The theory targets three fundamental questions:
\begin{description}
\item[\Qa.] What is the reason for a decision on an instance? 
\item[\Qb.] What minimal aspects of an instance will guarantee the decision on that instance?
\item[\Qc.] What minimal changes to an instance will lead to a different decision on that instance?
\end{description}

The answer to the first question will be a necessary and sufficient condition for the decision, expressed as a logical formula,
which captures the essence of the instance that led to the decision. 
The answer to the second question will also be a logical formula. It will characterize minimal sets of features with weakest 
conditions on their states that are guaranteed to trigger the decision. This will immediately identify features that are irrelevant to
the decision and will also identify irrelevant states of relevant features. The answer to the third question will also be a logical
formula but one that identifies minimal sets of features and how they may be changed to flip the decision into some other, designated or arbitrary, decision.
As we shall see later, the answer to the first question will form the basis for answering the second and third questions.

The discussed theory is based on representing classifiers using \textit{class formulas} which are logical formulas that
characterize the instances in each class. Two questions arise here. First, where do we get
these class formulas from? Second, is it always possible to represent a classifier this way? The first question may arise
when considering a classifier such as the one in Fig.~\ref{fig:nbc} which is numeric in nature and is based on probabilistic
reasoning. The features of this classifier are discrete so the instances in each class are finite and can be
represented by a logical formula that is obtained through a \textit{compilation} process to be discussed
at the end of this tutorial, in Section~\ref{sec:compilation}. The second question arises in the context of classifiers
like the decision tree in Fig.~\ref{fig:dtree-c} which involves an infinite number of instances due to the presence of
numeric features. It is well known that such decision trees can be easily discretized as shown in
Fig.~\ref{fig:dtree-d}. Here, the numeric values of the feature $\Age$ are partitioned into three equivalence intervals (labeled $a_1, a_2, a_3$) 
and the ones for $\BMI$ are partitioned into four equivalence intervals (labeled $b_1, b_2, b_3, b_4$). 
Two instances whose features have point values in the same corresponding
intervals are isomorphic from the viewpoint of this classifier, leading to a finite number of equivalence classes for the instances in each class
so they can also be represented using logical formulas.
The same applies to a broad 
set of classifiers including decision graphs and random forests with majority voting. 
Again, the compilation of class formulas for such classifiers will be discussed in Section~\ref{sec:compilation}.

We start this tutorial by some technical preliminaries in Section~\ref{sec:prelim} and follow by addressing the three explainability 
questions in Sections~\ref{sec:reasons},~\ref{sec:sreasons} and~\ref{sec:nreasons}. We will then discuss the compilation of classifiers 
into class formulas in Section~\ref{sec:compilation} and finally close with some remarks in Section~\ref{sec:conclusion}.

\section{Discrete Logic}
\label{sec:prelim}

It is quite common in the literature to employ Boolean logic even in the presence of discrete variables which are treated by ``binarization.'' 
For example, a discrete variable with $n$ values may be encoded using $n$ Boolean variables; see~\cite{corr/abs-2007-01493} for a review of several encoding schemes. 
We will not follow this tradition in this tutorial. Instead, we will work with what one may call \textit{discrete logic} as done in~\cite{DarwicheJi22,corr/abs-2304.14760}.
In this logic, discrete variables are first class citizens that include Boolean variables as a special case. This is \textit{very critical} semantically for the discussed theory
of explainability and can be critical for computation too.\footnote{We have seen, for example, random forests whose discretization leads to
variables with thousands of values. We want to avoid binarization in this case.}
The bias towards Boolean logic is  due to its potent and vast computational machinery, developed over decades,
which does not apply directly to discrete variables. 
As we shall see, however, we now have potent tools for working with discrete variables in the context of explainable AI.

\subsection{Syntax and Semantics of Discrete Logic}
We assume a finite set of discrete variables \(\Sigma\).
Each variable \(X \in \Sigma\) has a finite number of 
\textit{states} \(x_1, \ldots, x_n,\) \(n > 1\). 
A \textit{literal} \(\lit\) for variable \(X\), called an \(X\)-literal,
is a set of states such that \(\emptyset \subset \lit \subset \{x_1, \ldots, x_n\}\). 
We often denote a literal such as \(\{x_1,x_3,x_4\}\) 
by \(x_{134}\) which reads: the state of variable \(X\) 
is either \(x_1\) or \(x_3\) or \(x_4\).
A literal is \textit{simple} iff it contains
a single state. Hence, \(x_3\) is a simple literal but
\(x_{134}\) is not.
Since a simple literal corresponds to a state, 
we use these two notions interchangeably. 
If a variable \(X\) has two states, it is said to be \textit{Boolean} or  \textit{binary}
and its states are denoted by $x$ and $\NOT x$.
Such a variable can only have two simple literals, \(\{x\}\) and \(\{\NOT{x}\}\),
denoted \(x\) and \(\NOT x\) for convenience. 
A \textit{formula} is either a constant \(\top\), \(\bot\),
literal \(\lit\), negation \(\NOT \alpha\), conjunction
\(\alpha \AND \beta\) or disjunction \(\alpha \OR \beta\) 
where \(\alpha\), \(\beta\) are formulas. 
The set of variables appearing in a formula $\Delta$
are denoted by $\vars(\Delta)$.

The semantics
of discrete logic directly generalize those of Boolean
logic. A \textit{world,} denoted \(\w\), maps each variable in \(\Sigma\) to
one of its states.
A world \(\w\) is called a \textit{model} of formula \(\alpha\),
written \(\w \models \alpha\), iff \(\alpha\) is satisfied 
by \(\w\) (that is, \(\alpha\) is true at \(\w\)). 
The constant \(\top\) denotes a valid formula (satisfied by every world) 
and the constant \(\bot\) denotes an unsatisfiable formula (has no models). 
Formula \(\alpha\) implies/entails formula \(\beta\), 
written \(\alpha \models \beta\), iff
every model of \(\alpha\) is also a model of \(\beta\).
In this case, we say  $\alpha$ is \textit{stronger} than $\beta$ and also say $\beta$ is \textit{weaker} than $\alpha$.

\subsection{Representing Classifiers using Class Formulas}
A discrete variable \(X \in \Sigma\) is called a \textit{feature} and 
a  state \(x_i\) (or simple literal $\{x_i\}$) is called a \textit{characteristic.}
An \textit{instance} is a conjunction of characteristics, one for each feature in \(\Sigma\) (instances are in one-to-one correspondence with worlds).
\begin{definition}\label{def:classifier}
A \textit{classifier} with classes $c_1, \ldots, c_n$ is a set of mutually 
exclusive and exhaustive formulas \(\Delta_1, \ldots, \Delta_n\), called \textit{class formulas.} 
Instance \(\inst\) \textit{is in class} $c_i$ iff \(\inst \models \Delta_i\).
\end{definition}

The decision graph below is a classifier with three ternary features \(X, Y, Z\)
and three classes $c_1, c_2, c_3$.
Its class formulas are
$\Delta_1 = x_{12}\OR x_3 \AND y_1 \AND z_{13}$, 
$\Delta_2 = x_3 \AND z_2$ and
$\Delta_3 = x_3 \AND y_{23} \AND z_{13}$. 
\begin{wrapfigure}[8]{r}{0.175\textwidth}
\centering
 \scalebox{0.75}{
        \begin{tikzpicture}[
        roundnode/.style={circle ,draw=black, thick},
        squarednode/.style={rectangle, draw=black, thick},
        ]
        \node[squarednode]     (X)                              {\Large X};
        \node[roundnode]     (c1)       [below=of X, xshift = -0.8cm, yshift = 0.5cm] {\Large $c_1$};
        \node[squarednode]     (Y)       [below=of X, xshift = 0.8cm, yshift = 0.5cm] {\Large $Y$};
        \node[squarednode]       (Z1)   [below=of Y, xshift = -0.8cm, yshift = 0.5cm] {\Large $Z$};
        \node[squarednode]       (Z2)   [below=of Y, xshift = 0.8cm, yshift = 0.5cm] {\Large $Z$};
        \node[roundnode]       (c2)   [below=of Z1, xshift = -0.8cm, yshift = 0.5cm] {\Large $c_1$};
        \node[roundnode]       (c3)   [below=of Z1, xshift = 0.8cm, yshift = 0.5cm] {\Large $c_2$};
        \node[roundnode]       (c4)   [below=of Z2, xshift = 0.8cm, yshift = 0.5cm] {\Large $c_3$};
        
        \draw[-latex, thick] (X.240) -- node [anchor = center, xshift = -4mm, yshift = 1mm] {$x_1x_2$} (c1.north);
        \draw[-latex, thick] (X.300) -- node [anchor = center, xshift = 4mm, yshift = 1mm] {$x_3$} (Y.north);
        \draw[-latex, thick] (Y.240) -- node [anchor = center, xshift = -4mm, yshift = 1mm] {$y_1$} (Z1.north);
        \draw[-latex, thick] (Y.300) -- node [anchor = center, xshift = 4mm, yshift = 1mm] {$y_2y_3$} (Z2.north);
        \draw[-latex, thick] (Z1.240) -- node [anchor = center, xshift = -4mm, yshift = 1mm] {$z_1z_3$} (c2.north);
        \draw[-latex, thick] (Z1.300) -- node [anchor = center, xshift = -4mm] {$z_2$} (c3.north);
        \draw[-latex, thick] (Z2.240) -- node [anchor = center, xshift = 4mm] {$z_2$} (c3.north);
        \draw[-latex, thick] (Z2.300) -- node [anchor = center, xshift = 4mm, yshift = 1mm] {$z_1z_3$} (c4.north);
        \end{tikzpicture}
        }
\end{wrapfigure}
This classifier has \(27\) instances:
\(20\) in class \(c_1\), \(3\) in class \(c_2\)
and \(4\) in class \(c_3\). For example,
instance $\inst = x_3 \AND y_2 \AND z_2$ is
in class \(c_2\) since $\inst \models \Delta_2$.
Again, even though a classifier may take different forms, the presented theory of
explainability views it as a set of class formulas since this is all we need
to explain decisions.

\subsection{Explanations as Normal Forms}
The answers to questions \Qa, \Qb\ and \Qc\ will be expressed using logical formulas that have normal forms, discussed next.
A \textit{term} is a conjunction of literals for distinct variables. 
A \textit{clause} is a disjunction of literals for distinct variables.
A \textit{Disjunctive Normal Form (DNF)} is a disjunction of terms.
A \textit{Conjunctive Normal Form (CNF)} is a conjunction of clauses.
A \textit{Negation Normal Form (NNF)} is a formula with no negations.
These definitions imply that terms cannot be inconsistent,
clauses cannot be valid, DNFs and CNFs are NNFs, and none
of these normal forms can include negations.  
We say a formula (term/clause) is \textit{simple} if it contains only simple literals.
Simple and non-simple formulas will delineate two approaches
for answering questions \Qa, \Qb\ and \Qc\ based on the amount of information they convey about decisions.

\subsection{Constructing Abstractions using Conditioning}
When constructing abstractions of instances to explain decisions, the notion of
conditioning plays a central role. Observe first that a simple term like $x_1 \AND y_3 \AND z_2$ represents a set of states.
The \textit{conditioning} of formula \(\Delta\) on a simple term \(\tau\)
is denoted by \(\Delta | \tau\) and obtained as follows. 
For each state \(x_i\) that appears in the simple term \(\tau\),  
replace each \(X\)-literal \(\lit\) in \(\Delta\) with \(\top\) 
if \(x_i \in \lit\); otherwise, replace \(\lit\) with \(\bot\).
Consider the formula \(\Delta = x_{12} \OR x_3 \AND y_1 \AND z_{13}\) and the simple term \(\tau = x_3 \AND z_1\).
Then \(\Delta | \tau = \bot \OR \top \AND y_1 \AND \top = y_1\).
In general, the conditioned formula \(\Delta | \tau\) does not mention any variable that appears in term \(\tau\).

\subsection{Minimality using Prime Implicants and Implicates}
Questions \Qb\ and \Qc\ involve a notion of minimality that will be captured using the classical notions of
prime implicants and implicates. 
A term \(\tau\) is called an \textit{implicant} of formula $\Delta$ iff it implies \(\Delta\) (\(\tau \models \Delta\)). 
It is called a \textit{prime implicant} if no other implicant of \(\Delta\) is weaker than \(\tau\) (i.e.,~\(\tau \models \tau' \models \Delta\) does not hold for any term \(\tau' \neq \tau\)).
A clause \(\sigma\) is an \textit{implicate} of formula $\Delta$ iff it is implied by \(\Delta\) ($\Delta \models \sigma$).
It is called a \textit{prime implicate} if no other implicate of \(\Delta\) is stronger than \(\sigma\) (i.e.,~$\Delta \models \sigma' \models \sigma$ does not hold for any clause \(\sigma' \neq \sigma\)).
A prime implicant \(\tau\) is \textit{variable-minimal} iff no other prime implicant \(\tau'\) satisfies \(\vars(\tau') \subset \vars(\tau)\).
Variable-minimal prime implicates are defined similarly.

The prime implicants and implicates of discrete formulas behave in ways that may surprise someone
who is accustomed to these notions in a Boolean setting. Moreover, these behaviors have interesting
implications on questions \Qb\ and \Qc. Consider the 
Boolean formula \(\Delta_b = (x \OR \NOT{z}) \AND (x \AND z \OR y)\). The term $x \AND y \AND \NOT{z}$ is an implicant but not prime. 
The only way to weaken this term so it becomes prime is by dropping some of its variables ($y \AND \NOT{z}$ is a prime implicant). 
In discrete logic, we can possibly weaken a term without dropping any of
its variables, by adding states to some of its literals. 
Consider the discrete formula \(\Delta_d = (x \AND z_1 \OR z_2) \AND (x \AND z_3 \OR y)\), where variable $Z$ is ternary and variables $X, Y$ are binary.
The term $x \AND y \AND z_1$ is an implicant of this formula but is not prime. We can weaken
this term by adding the state $z_2$ to literal $z_1$, leading to the prime implicant  $x \AND y \AND z_{12}$.
A symmetrical situation arises for prime implicates: the only way to strengthen a Boolean clause is by dropping some of its
variables but we can strengthen a discrete clause by removing states from its literals. 

\section{The Reasons Behind Decisions}
\label{sec:reasons}

We now turn to answering question \Qa: What is the reason for a decision on an instance? We will present two different
answers to this question. The first underlies extensive developments in explainable AI over the last few years. The second
is very recent, more informative and further subsumes the first answer. The two answers, however, are equivalent if all
features of the classifier are binary (i.e., Boolean).

Consider a classifier specified by class formulas \(\Delta_1, \ldots, \Delta_n\) and suppose it decides that
instance \(\inst\) belongs to class \(c_i\), \(\inst \models \Delta_i\). We need to know why. The answer to this question
must be a condition on the instance that implies formula \(\Delta_i\). Since this condition is meant to represent
an abstraction of the instance, we want it to be as weak as possible so we get the most general abstraction. 
Here where things get interesting.  
If we do not place additional requirements on this condition, we will get a trivial answer to question~\Qa\ since the 
class formula \(\Delta_i\) satisfies the requirements we stated. To see why, \(\inst \models \Delta_i\) by supposition
so \(\Delta_i\) is indeed a condition satisfied by the instance. Moreover, there is no other condition on instance \(\inst\)
that is weaker than \(\Delta_i\) and that implies \(\Delta_i\). This answer is trivial as it will be the same answer
returned when explaining the decision on any instance in class \(c_i\). 
Hence, we need an additional requirement on the sought condition and we next review two proposals
that have been extended for this purpose.

\subsection{Complete Reasons}

The first proposal requires the sought condition---that is, the reason for the decision---to be constructed
from states that appear in the instance which can be combined in any fashion but using 
only conjunctions and disjunctions. Technically speaking, it requires the condition to be a simple NNF formula
whose literals all appear in the instance being explained. This leads to the notion of \textit{complete
reason} introduced in~\cite{ecai/DarwicheH20}.\footnote{Ref.~\cite{ecai/DarwicheH20} gave a different but equivalent definition
and for binary variables. See~\cite{DarwicheJi22} for a more general treatment.}

\begin{definition}\label{def:complete reason}
Suppose instance \(\inst\) belongs to class \(c_i\). 
The \textit{complete reason} for the decision on instance \(\inst\) is 
the weakest NNF $\Gamma$ whose literals are in $\inst$
and that satisfies $\inst \models \Gamma \models \Delta_i$.
\end{definition}

We stress here that the complete reason is a simple NNF formula, that is, every literal it contains
corresponds to a state.

Consider a classifier with
three ternary features \(X, Y, Z\). Instance \(\inst = x_2 \AND y_2 \AND z_1\) is
decided as belonging to class \(c_2\) which has formula \(\Delta_2 = x_{23} \AND (x_2 \OR y_{23}) \AND (y_{23} + z_1)\).
The complete reason for this decision is \(x_2 \AND (y_2 \OR z_1)\). It says that instance \(\inst\)
belongs to class \(c_1\) because it has characteristic \(x_2\) and one of the two characteristics \(y_2, z_1\). 
This is indeed the most general condition on the instance which implies the class formula and that
is constructed by applying conjunctions and disjunctions to the instance characteristics $x_2$, $y_2$ and $z_1$.

Two questions become relevant now. First, how do we compute the complete reason? Second, how can we use
complete reasons when they are too complex to be interpretable by humans? We will address the first question next
while leaving the second question until later in the tutorial. 

Ref.~\cite{ecai/DarwicheH20} showed that if the class formula is appropriately represented (e.g., as an OBDD~\cite{tc/Bryant86}),
the complete reason can be computed in linear time. A major development then came in~\cite{jair/DarwicheM21} which
introduced the following quantification operator and showed that it can be used to express the complete reason.

\begin{definition}\label{def:forall}
For variable \(X\) with states \(x_1, \ldots, x_n\),
the \textit{universal literal quantification} of state $x_i$ from formula $\Delta$ is defined 
as $\forall x_i \cdot \Delta = \Delta | x_i \AND \BAND_{j \not = i} (x_i \OR \Delta | x_j)$.
\end{definition}

This operator is commutative so it is meaningful to quantify an instance \(\inst\) from its
class formula \(\Delta_i,\) written \(\forall \inst \cdot \Delta_i\), by quantifying the states of \(\inst\) in any order. 
Moreover, \(\forall \inst \cdot \Delta_i\) corresponds to the complete reason as shown in~\cite{jair/DarwicheM21} for Boolean variables 
and in~\cite{DarwicheJi22} for discrete variables. One significance of this result is that
this quantification operator is well behaved computationally on a number of logical forms. For example,
one can quantify any set of states from a CNF in linear time.
More generally, the operator distributes over conjunctions, and distributes over disjunctions when the
disjuncts do not share variables as also shown in~\cite{jair/DarwicheM21} (see weaker conditions in~\cite{DarwicheJi22}).
This brought into focus the following form of class formulas as it allows one to compute complete reasons in linear time (and general reasons too
that are discussed in Section~\ref{sec:general reasons}).
\begin{definition}\label{def:or-decomposable}
An NNF is \textit{or-decomposable} iff \(\vars(\alpha)\cap\vars(\beta)=\emptyset\) for any disjuncts \(\alpha\) and \(\beta\) in the NNF.\footnote{This is to
be contrasted with and-decomposable NNFs, known as Decomposable NNFs (DNNFs)~\cite{jacm/Darwiche01}, which have been studied extensively.}
\end{definition}

Due to this form and some of its weakenings,~\cite{DarwicheJi22}
provided closed-form complete reasons for certain classes of classifiers that are based on decision graphs.

\subsection{General Reasons}
\label{sec:general reasons}

We now turn to a recent development~\cite{corr/abs-2304.14760} 
that identified a weaker requirement on abstractions of instances which can produce more 
information when explaining decisions.\footnote{Def.~\ref{def:general reason} is actually a proposition in~\cite{corr/abs-2304.14760}.
We reversed the formulation here to serve the storyline in this tutorial.}

\begin{definition}\label{def:general reason}
Suppose instance \(\inst\) is in class \(c_i\). 
The \textit{general reason} of the decision on instance \(\inst\) is the weakest NNF \(\Gamma\) whose literals are
implied by \(\inst\) and that satisfies \(\inst \models \Gamma \models \Delta_i\).
\end{definition}

The only difference between the complete reason and the general reason is that the former requires literals to appear in
the instance, while the latter only requires that they are implied by the instance. Hence, general reasons are
no longer simple formulas like complete reasons as they may contain non-simple literals. If all features are binary, then every
literal is simple and the complete and general reasons are equal. However, in the presence of non-binary features,
general reasons provide more information about why a decision was made. Let us see this through an example.

\begin{figure}[tb]
\centering
        \scalebox{0.75}{
        \begin{tikzpicture}[
        roundnode/.style={text width = 0.55cm, circle ,draw=black, thick, text badly centered},
        squarednode/.style={rectangle, draw=black, thick, text badly centered},
        ]
        \node[squarednode]      (Age)                              {\large \Db};
        \node[squarednode]        (Weight)       [below=of Age, xshift = -0.9cm, yshift = 0.4cm] {\large \Weight};
        \node[roundnode]        (No1)       [below=of Age, xshift = 0.9cm, yshift = 0.5cm] {\large \no};
        \node[roundnode]        (Yes1)       [below=of Weight, xshift = -2.3cm, yshift = 0.2cm] {\large \yes};
        \node[squarednode]        (BloodType1)       [below=of Weight, yshift = 0cm] {\large \BType};
        \node[squarednode]        (BloodType2)       [below=of Weight, xshift = 2.5cm, yshift = 0cm] {\large \BType};
        \node[roundnode] (Yes2) [below=of BloodType1, xshift = -1.3cm, yshift = 0.3cm] {\large \yes};
        \node[roundnode] (No2) [below=of BloodType1, xshift = 0cm, yshift = 0.3cm] {\large \no};
        \node[roundnode] (Yes3) [below=of BloodType2, xshift = -1cm, yshift = 0.3cm] {\large \yes};
        \node[roundnode] (No3) [below=of BloodType2, xshift = 0.5cm, yshift = 0.3cm] {\large \no};
        
        \draw[-latex, thick] (Age.240) -- node [anchor = center, xshift = -5mm, yshift = 1mm] {$\yy$} (Weight.north);
        \draw[-latex, thick] (Age.300) -- node [anchor = center, xshift = 5mm, yshift = 1mm] {$\nn$} (No1.north);
        \draw[-latex, thick] (Weight.240) --  node [anchor = center, xshift = -12mm] {\OWeight} (Yes1.north);
        \draw[-latex, thick] (Weight.270) --  node [anchor = center, xshift = -0mm] {\UWeight} (BloodType1.north);
        \draw[-latex, thick] (Weight.300) --  node [anchor = center, xshift = 10mm] {\Nom} (BloodType2.north);
        \draw[-latex, thick] (BloodType1.240) --  node [anchor = center, xshift = -9mm] {\tA, \tB, \tAB} (Yes2.north);
        \draw[-latex, thick] (BloodType1.270) --  node [anchor = center, xshift = 3mm] {\tO} (No2.north);
        \draw[-latex, thick] (BloodType2.240) --  node [anchor = center, xshift = -6mm] {\tA, \tB} (Yes3.north);
        \draw[-latex, thick] (BloodType2.270) --  node [anchor = center, xshift = 6mm] {\tAB, \tO} (No3.north);
        \end{tikzpicture}
        }
\caption{A classifier of some disease in the form of a decision tree~\cite{corr/abs-2304.14760}.}
\label{fig:disease-dgraph}
\end{figure}
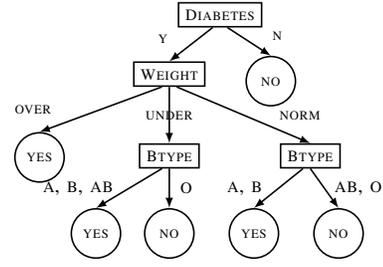

Consider the classifier in Fig.~\ref{fig:disease-dgraph} and a patient Rob,
\[(\eql{\Db}{\yy}) \AND (\eql{\Weight}{\UWeight}) \AND (\eql{\BType}{\tA}),\]
decided as $\yes$.
The complete reason for the decision is
\[(\eql{\Db}{\yy}) \AND (\eql{\BType}{\tA})\]
and the general reason is
\begin{align*}
& (\eql{\Db}{\yy}) \AND (\BType\IN\{\tA,\tB,\tAB\}) \AND \\
&~~~~(\Weight\IN\{\UWeight,\OWeight\} \OR \BType\IN\{\tA,\tB\}).
\end{align*}
The complete reason justifies the decision by Rob
having $\Db$ and a $\BType$ of $\tA$. But the general reason provides weaker justifications. 
One of them is that Rob has $\Db$ and his $\BType$ is not $\tAB$ or $\tO$.
Another justification is that he has $\Db$, his $\Weight$ is not $\Nom$ and his $\BType$ is not $\tO$.

The additional information provided by general reasons is particularly critical in the context of discretized classifiers.
Consider again the decision tree in Fig.~\ref{fig:decision-tree} which has two numeric features $\Age$, $\BMI$.
The feature $\Age$ is discretized into three intervals $[0, 18), [18, 40)$ and $[40, \infty)$ with corresponding labels \(a_1, a_2, a_3\).
The feature $\BMI$ is discretized into four intervals $[0, 25), [25, 27), [27, 30), [30, \infty)$ with corresponding labels $b_1, b_2, b_3, b_4$.
Hence,  $\Age$ can be treated as a discrete variable with states \(a_1, a_2, a_3\) and $\BMI$ can be treated as a discrete variable with
states $b_1, b_2, b_3, b_4$. 
Consider now the instance $\eql{\Age}{42} \AND \eql{\BMI}{28}$ which is
discretized into $a_3 \AND b_3$ ($\Age \geq 40$ and $27 \leq \BMI < 30$).
The decision on this instance is $\yes$ and 
its class formula \(\Delta_{\yes}$ is $a_1 \AND b_4 \OR a_2 \AND b_{34} \OR a_3 \AND b_{234}\) (obtained
by tracing the paths leading into class $\yes$). 
The complete reason for this decision is $a_3 \AND b_3$ which is the instance itself
so no abstraction took place.
The general reason is $a_{23}\AND b_{34} \OR a_3 \AND b_{234}$ which reads
$$(\Age \ge 18) \AND (\BMI \ge 27) \OR
 (\Age \ge 40) \AND (\BMI \ge 25).$$ 
This is a weaker property of the instance that still implies the decision.
For example, the first part says the decision is justified by  \(\Age\) being no less than \(18\) 
and \(\BMI\) being no less than \(27\).
These kind of justifications are impossible to obtain using complete reasons as such reasons can only reference 
instance characteristics, that is, $\Age \geq 40$ and $27 \leq \BMI < 30$.

General reasons can also be expressed using a quantification operator introduced in~\cite{corr/abs-2304.14760}.
\begin{definition}\label{def:iuq}
For variable \(X\) with states \(x_1, \ldots, x_n\) and formula \(\Delta\),
$\iuq x_i \cdot \Delta$ is defined as $\Delta | x_i \AND \Delta$.
\end{definition}

This operator is also commutative so we can quantify an instance \(\inst\) from its
class formula \(\Delta_i,\) written \(\iuq \inst \cdot \Delta_i\), by quantifying the states of \(\inst\) in any order. 
Moreover,  \(\iuq \inst \cdot \Delta_i\) corresponds to the general reason for the decision on instance \(\inst\)
as shown in~\cite{corr/abs-2304.14760}. The operators \(\forall\) and \(\iuq\) have similar computational properties
as far as distributivity and  tractability on forms such as CNFs and or-decomposable NNFs. This is why,
similar to complete reasons,
we also have closed-form general reasons for certain classes of classifiers~\cite{corr/abs-2304.14760}.

\subsection{Fixation and Monotonocity}

Complete and general reasons satisfy a property called fixation, which has key implications including on computation
as we discuss later. This property was also identified in~\cite{corr/abs-2304.14760}.

\begin{definition}\label{def:fixation}
An NNF is \textit{locally fixated} on instance $\inst$  iff its literals are consistent with~$\inst$.
A formula is \textit{fixated} on instance $\inst$ iff it is equivalent to an NNF that is locally fixated on $\inst$.
\end{definition}

The formula $x_{12} \AND y_{13} \OR z_{23}$ is locally fixated on the instance $\inst = x_1\AND y_1 \AND z_2$,
but this is not true for the formula $x_{12} \AND y_{23} \OR z_{23}$ since the literal $y_{23}$ is not consistent
with the instance $\inst$.
The fixation of complete reasons follows directly from Def.~\ref{def:complete reason} and the fixation of 
general reasons follows directly from Def.~\ref{def:general reason}.
 
If a simple formula is fixated then it is also monotone.

\begin{definition}\label{def:monotonicity}
A formula is \textit{monotone} iff for each variable $X$ all $X$-literals that occur in the formula are simple and equal.
\end{definition}

For example, the formula $(x_1 \OR y_2) \AND (x_1 \OR z_3) \AND (y_2 \OR z_3)$ is monotone,
but the formulas $(x_1 \OR y_2) \AND (x_{12} \OR z_3) \AND (y_2 \OR z_3)$ is not monotone
since it contains a non-simple literal $x_{12}$. 
The formula $(x_1 \OR y_2) \AND (x_1 \OR z_3) \AND (y_3 \OR z_3)$ is not monotone either 
since it contains distinct literals $y_2$ and $y_3$ for variable $Y$.
Complete reasons are known to be monotone and this property was exploited 
computationally in~\cite{ecai/DarwicheH20,DarwicheJi22} when computing the prime
implicants and implicates of complete reasons.

\subsection{Selection Semantics}

For an instance \(\inst\) and its class formula \(\Delta_i\), we have~\cite{corr/abs-2304.14760}:
\[ \inst \models \forall \inst \cdot \Delta_i \models \iuq \inst \cdot \Delta_i \models \Delta_i \]
so the general reason is always weaker than the complete reason. 
The above relations also show that reasons, whether complete or general,
are stronger than class formulas. Hence, the operators \(\forall\) and $\iuq$ can be
viewed as \textit{selection operators} since they select subsets of the instances in 
class $c_i$~\cite{jair/DarwicheM21,corr/abs-2304.14760}. 
In particular,
$\forall \inst \cdot \Delta_i$ selects all instances in class
\(c_i\) whose membership in the class does not depend on 
characteristics that are inconsistent with instance \(\inst\).
For example, suppose that instance $\inst'$ is in class $c_i$ and 
disagrees with instance $\inst$ only on the states of features $X, Y, Z$.
Then $\inst'$ will be selected by $\forall \inst \cdot \Delta_i$ iff
changing the states of any subset of its features $X, Y, Z$ yields an instance that is also in class $c_i$.
In contrast, $\iuq \inst \cdot \Delta_i$ selects all instances
that remain in class $c_i$ if any of their characteristics are changed to agree 
with instance $\inst$ (these include the ones selected by $\forall \inst \cdot \Delta_i$). In the
previous example, instance $\inst'$ will be selected by $\iuq \inst \cdot \Delta_i$ iff 
setting the states of any subset of its features $X, Y, Z$ to the states they have in $\inst$ yields an instance
that is also in class $c_i$.

\section{The Sufficient Reasons for a Decision}
\label{sec:sreasons}

We now turn to answering question~\Qb: What minimal aspects of an instance guarantee the decision on that instance?
We will provide two answers which depend on how we define ``aspects.''
The first answer is based on the following notion.

\begin{definition}\label{def:sr}
Suppose instance \(\inst\) is in class \(c_i\). 
The \textit{sufficient reasons (SRs)} for the decision on instance \(\inst\) are the prime implicants of the complete reason $\forall \inst \cdot \Delta_i$.
\end{definition}

A sufficient reason is a term \(\tau = \lit_1 \AND \ldots \AND \lit_n\) where each literal \(\lit_i\) is simple (i.e., a state) and appears in 
instance \(\inst\).\footnote{Since the complete reason is simple and fixated on the instance \(\inst\).}
This leads to the relationships \(\inst \models \tau \models \forall \inst \cdot \Delta_i \models \Delta_i\)
which further imply that any instance containing states  $\lit_1, \ldots, \lit_n$ will be decided similarly to instance \(\inst\). Moreover,
this does not hold for any strict subset of these states; otherwise, $\tau$ cannot be a prime implicant of $\forall \inst \cdot \Delta_i$. 
In other words, the sufficient reason \(\tau\) is a minimal set of characteristics in instance \(\inst\)
that justify the decision on the instance, so other characteristics of the instance can be viewed as irrelevant to the
decision as they can be changed in any fashion while sustaining the decision. 

The notion of a sufficient reason was first introduced in~\cite{ijcai/ShihCD18} under the name of a \textit{PI-explanation} 
using a different but equivalent definition.
The term ``sufficient reason'' was first used in~\cite{ecai/DarwicheH20} which introduced the 
complete reason and showed that its prime implicants are the PI-explanations.
Sufficient reasons are sometimes also called \textit{abductive explanations}~\cite{IgnatievNM19a}.

\begin{figure}[tb]
\centering
\includegraphics[width=6.0cm]{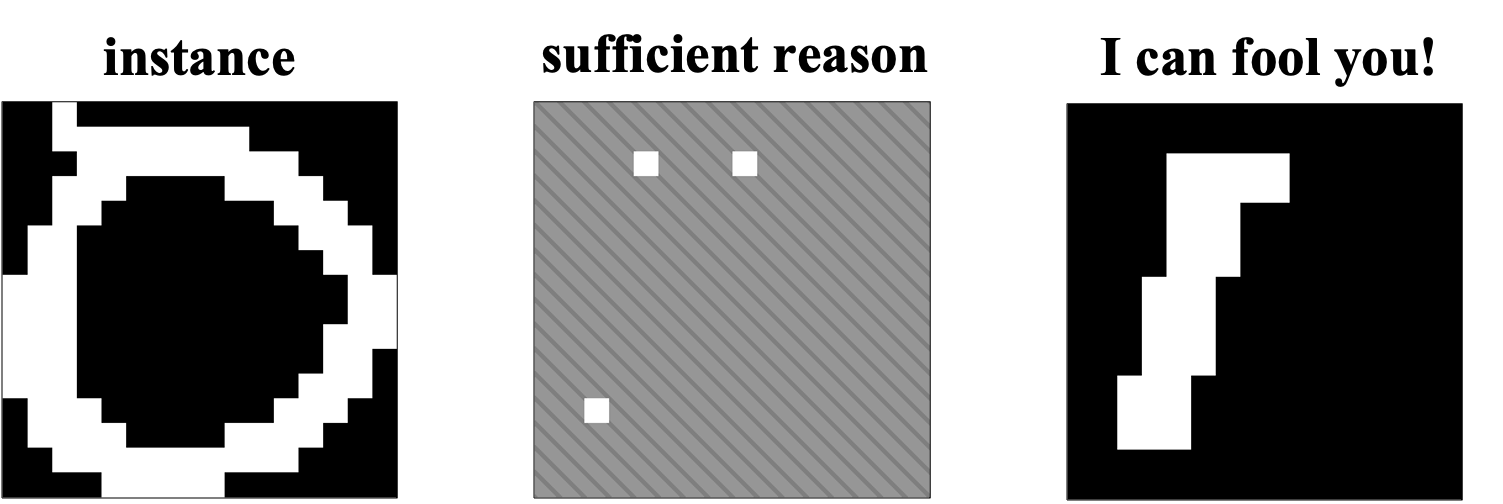}
\caption{Example sufficient reason from the domain of classifying digits~\cite{kr/ShiSDC20}.}
\label{fig:digits}
\end{figure}

For a concrete example of sufficient reasons, from~\cite{kr/ShiSDC20}, consider Fig.~\ref{fig:digits}
which depicts $16\mathord\times\mathord16$ images. The image on the left was passed to a
Convolutional Neural Network (CNN) that is tasked with classifying digits~$0, 1$ and
the CNN classified the image correctly as digit~$0$. 
One of the sufficient reasons for this decision is depicted in the middle of Fig.~\ref{fig:digits}.
This sufficient reason contains~$3$ white pixels (out of~$256$ pixels) so any image that contains
these $3$ white pixels will be classified as digit~$0$ by this CNN, as shown on the
right of Fig.~\ref{fig:digits}.

For another concrete example, from~\cite{corr/abs-2304.14760}, 
consider the classifier in Fig.~\ref{fig:disease-dgraph} and the following patient, Lara,
\begin{equation}
\inst\mathord:~~(\eql{\Db}{\yy}) \AND (\eql{\Weight}{\OWeight}) \AND (\eql{\BType}{\tA}). \label{lara}
\end{equation}
The decision on Lara is $\yes$ and the class formula $\Delta_{\yes}$ is
\begin{align}
& (\eql{\Db}{\yy}) \AND \nonumber \\
&~~~[(\eql{\Weight}{\OWeight})\ \OR \nonumber \\
&~~~~(\eql{\Weight}{\Nom}) \AND (\BType\IN\{\tA,\tB\})\ \OR \nonumber \\
&~~~~(\eql{\Weight}{\UWeight}) \AND (\BType\IN\{\tA,\tB,\tAB\})]. \label{cf:disease}
\end{align}
The complete reason $\forall \inst \cdot \Delta_{\yes}$ for this decision is
\begin{equation}
 (\eql{\Db}{\yy}) \AND [(\eql{\Weight}{\OWeight}) \OR (\eql{\BType}{\tA})]. \label{cr:disease}
\end{equation}
It has two prime implicants
\begin{align}
& (\eql{\Db}{\yy}) \AND (\eql{\Weight}{\OWeight}) \label{sr1:lara} \\
& (\eql{\Db}{\yy}) \AND (\eql{\BType}{\tA}) \label{sr2:lara}
\end{align}
which are the sufficient reasons for the decision. Hence, the feature
$\BType$ is irrelevant to the decision on Lara given her characteristics $(\eql{\Db}{\yy}) \AND (\eql{\Weight}{\OWeight})$,
and the feature $\Weight$ is irrelevant given  $(\eql{\Db}{\yy}) \AND (\eql{\BType}{\tA})$.

One can obtain weaker sufficient reasons---and, hence, identify further irrelevant aspects of instances---by 
employing the general reason behind the decision as shown recently in~\cite{corr/abs-2304.14760}. 

\begin{definition}\label{def:gsr}
Suppose instance \(\inst\) is in class \(c_i\). 
The \textit{general sufficient reasons (GSRs)} for the decision on \(\inst\) are the variable-minimal prime implicants 
of  the general reason $\iuq \inst \cdot \Delta_i$.\footnote{This
definition is a proposition in~\cite{corr/abs-2304.14760} which provided a different definition for general sufficient reasons
that does not reference general reasons.}
\end{definition}

The general reason $\iuq \inst \cdot \Delta_{\yes}$ for the decision on Lara is
\begin{align}
& (\eql{\Db}{\yy}) \AND \nonumber \\
&~~~[(\eql{\Weight}{\OWeight}) \OR (\BType\IN\{\tA,\tB\})\ \OR  \nonumber \\
&~~~~(\Weight\IN\{\UWeight,\OWeight\}) \AND (\BType\IN\{\tA,\tB,\tAB\})] \label{gr:disease}
\end{align}
and has three prime implicants
\begin{align}
\hspace{-1mm}(\eql{\Db}{\yy}) & \AND (\eql{\Weight}{\OWeight}) \label{pi1:disease} \\
\hspace{-1mm}(\eql{\Db}{\yy}) & \AND (\BType\IN\{\tA, \tB\}) \label{pi2:disease} \\
\hspace{-1mm}(\eql{\Db}{\yy}) & \AND (\Weight\IN\{\UWeight,\OWeight\}) \AND (\BType\IN\{\tA,\tB,\tAB\}) \label{pi3:disease}
\end{align}
Only the first two are variable-minimal so they are the general sufficient reasons for the decision.
We have seen~(\ref{pi1:disease}) as a sufficient reason in~(\ref{sr1:lara}). 
However, the general sufficient reason in~(\ref{pi2:disease}) is weaker than the sufficient reason in~(\ref{sr2:lara})
and conveys more information about the decision. In particular, the sufficient reason in~(\ref{sr2:lara}) says that Lara's $\Weight$
is irrelevant to the decision because she has $\Db$ and her $\BType$ type is $\tA$, but the general sufficient reason  in~(\ref{pi2:disease})
says it is because she has $\Db$ and her $\BType$ is not $\tAB$ or $\tO$.

Every general sufficient reason is consistent with the instance since the general reason is fixated on the instance.
Define the \textit{intersection} of instance \(\inst\) with a general sufficient reason \(\tau\) as the smallest subset of \(\inst\) that implies \(\tau\).
Then this intersection is guaranteed to be a sufficient reason. Moreover, this guarantee holds iff we have 
the variable-minimality condition in Def.~\ref{def:gsr}. For example, 
the intersection of~(\ref{lara}) and~(\ref{pi2:disease}) is $(\eql{\Db}{\yy}) \AND (\eql{\BType}{\tA})$ which is a sufficient reason but
the intersection of~(\ref{lara}) and~(\ref{pi3:disease}) is $(\eql{\Db}{\yy}) \AND (\eql{\Weight}{\OWeight}) \AND (\eql{\BType}{\tA})$ 
which is not a sufficient reason.\footnote{Ref.~\cite{corr/abs-2007-01493} suggested using prime implicants that are not variable-minimal.}
More generally, the sufficient reasons of a decision are precisely the intersections of general sufficient reasons
with the instance. Hence, general sufficient reasons subsume their classical counterparts and provide more information about
decisions and underlying classifiers. We finally note, again, that if all features are binary, the complete and general reasons are equivalent so
general sufficient reasons reduce to classical sufficient reasons.
All the previous observations are implied by the results in~\cite{corr/abs-2304.14760}.

\section{The Necessary Reasons for a Decision}
\label{sec:nreasons}

We now turn to answering question \Qc: What minimal changes to an instance will lead to a different decision on that instance?
The answer to this question will identify minimal sets of features that will flip the decision if changed properly. There are two
orthogonal variations on this question. The first determines whether the specific changes to these features are also
identified which in turn depends on whether we employ the complete or general reason for this purpose. 
The second variation relates to whether we seek to flip
the decision into some specific alternative or some arbitrarily different decision.

\subsection{Undermining a Current Decision}

We assume in this section that our goal is to undermine the current decision, without targeting a particular alternate decision.
This aim can be achieved using the following notion.

\begin{definition}
Suppose instance \(\inst\) is in class \(c_i\). 
The \textit{necessary reasons (NRs)} for the decision on instance \(\inst\) are the prime implicates of the complete reason $\forall \inst \cdot \Delta_i$.
\end{definition}

Each necessary reason is a clause \(\lit_1 \OR \ldots \OR \lit_n\) where each literal \(\lit_i\) is simple (a state) and appears in instance \(\inst\).\footnote{Since
the complete reason is simple and fixated on the instance.} Hence, each necessary reason is satisfied by the instance.
However, if we change the instance in a way that violates a necessary reason, the decision is no longer guaranteed to be sustained.
Suppose state \(\lit_i\) is for feature \(X_i\). To violate the necessary reason \(\lit_1 \OR \ldots \OR \lit_n\) we must change the state of each
feature \(X_i\) in the instance. The guarantee that comes with a necessary reason is that at least one such change will flip the decision,
and no strict subset of this change will flip the decision.

The notion of a necessary reason was first formalized in~\cite{aiia/IgnatievNA020} using a different but equivalent definition, 
under the name of a \textit{contrastive explanation}~\cite{lipton_1990}. 
Necessary reasons are sometimes also called \textit{counterfactual explanations}~\cite{kr/AudemardKM20}.\footnote{There is an
extensive body of work in philosophy, social science and AI that discusses contrastive explanations and counterfactual
explanations; see, e.g.,~\cite{Garfinkel1982-GARFOE-3,Lewis1986-LEWCE,temple1988contrast,Counterfactual_blackbox,van2018contrastive,DBLP:journals/ai/Miller19,Wachter_Contrastive_2019,Counterfactual_Visual,verma2022counterfactual,Diverse_Counterfactual_Explanations}. 
While the definitions of these notions are sometimes variations or refinements on one another, they are not always compatible.}
The term ``necessary reason'' was first used in~\cite{DarwicheJi22}
which showed that contrastive explanations are the prime implicates of the complete reason and provided the semantics of
necessity discussed earlier (a property that must be preserved if the decision is to be preserved). Let us now illustrate necessary reasons
using a concrete example.

We are back to patient Lara given in~(\ref{lara}) who was decided as $\yes$ by the classifier in Fig.~\ref{fig:disease-dgraph}.
The class formula is given in~(\ref{cf:disease}) and 
the complete reason for the decision is given in~(\ref{cr:disease}).
This complete reason has two prime implicates, 
 $(\eql{\Db}{\yy})$ and $(\eql{\Weight}{\OWeight}) \OR (\eql{\BType}{\tA})$.
There is only one way to violate the first reason by setting $\Db$ to $\nn$, which will indeed flip the decision to \(\no\)
as can be verified in Fig.~\ref{fig:disease-dgraph}. The second necessary reason can be violated in six different ways,
by setting $\Weight$ to a state in $\{\UWeight, \Nom\}$ and setting $\BType$ to a state in $\{\tB,\tAB,\tO\}$.
Not all of these changes/violations will flip the decision but at least one will.
The 
change to $(\eql{\Weight}{\UWeight}) \AND (\eql{\BType}{\tO})$ does flip the decision but the 
change to $(\eql{\Weight}{\UWeight}) \AND (\eql{\BType}{\tAB})$ does not (always assuming that $\Db$ is left unchanged).
We stress again that no strict subset of the first change will flip the decision since the changes suggested by 
necessary reasons are minimal.

We will now see that we can do better than this if we employ the general reasons of decisions
as suggested recently in~\cite{corr/abs-2304.14760}.

\begin{definition}\label{def:gnr}
Suppose instance \(\inst\) is in class \(c_i\). 
The \textit{general necessary reasons (GNRs)} for the decision on  \(\inst\) are the variable-minimal prime implicates
of  general reason $\iuq \inst \cdot \Delta_i$.\footnote{This
definition is a proposition in~\cite{corr/abs-2304.14760} which provided a different definition for general necessary reasons
that does not reference general reasons.}
\end{definition}

Like necessary reasons, violating a general necessary reason will undermine the decision. However, we now
have an additional guarantee: \textit{every} violation of a general necessary reason will flip the decision.
This guarantee will not hold if we do not insist on variable-minimal prime implicates, and if some prime implicate
that is not variable minimal satisfies this guarantee, then some changes it suggests cannot be minimal (i.e., we
can flip the decision with fewer changes to the instance).
Moreover, any change
that flips the decision and is suggested by some necessary reason will also be suggested by some
general necessary reason. That is, the latter reasons subsume the former and convey more information
about decisions and the underlying classifiers.\footnote{The general necessary
reasons are fixated (Def.~\ref{def:fixation}) on instance \(\inst\).} 
Let us look at a concrete example.

The general reason for the decision on Lara is given in~(\ref{gr:disease})
and has three prime implicates
\begin{align*}
& (\eql{\Db}{\yy}) \\
& (\eql{\Weight}{\OWeight}) \OR (\BType\IN\{\tA,\tB,\tAB\}) \\
& (\Weight\IN\{\UWeight,\OWeight\}) \OR (\BType\IN\{\tA,\tB\}).
\end{align*}
All are variable-minimal so they are all general necessary reasons.
The first can be violated in only one way by setting $\Db$ to $\nn$ which will flip the decision.
The second can be violated in two ways by setting $\Weight$ to a state in $\{\UWeight,\Nom\}$
and setting $\BType$ to state $\tO$. Both violations will flip the decision as can be verified using
the classifier in Fig.~\ref{fig:disease-dgraph}.
The third general necessary reason can be violated in two ways by setting $\Weight$ to state $\Nom$
and setting $\BType$ to a state in $\{\tAB,\tO\}$. Again, both violations will flip the decision.
Hence, every violation of these general necessary reasons is guaranteed to flip the decision.
Moreover, every change suggested by a necessary reason and which flips the decision is also suggested by
some general necessary reason.

We conclude this section by the following remarks.
Boolean resolution derives the clause \(\alpha \OR \beta\) from clauses \(x \OR \alpha\) and \(\NOT{x} \OR \beta\).
Moreover, a classical method for computing prime implicates of Boolean CNFs is to close the
CNF under resolution and remove subsumed clauses after each resolution step; see, e.g.,~\cite{gurvich1999generating,Crama2011BooleanF}. 
As recently shown in~\cite{corr/abs-2304.14760},
this can also be done for discrete CNFs but using a generalized
resolution rule that derives clause $\lit \AND \lit' \OR \alpha \OR \beta$ from clauses $\lit \OR \alpha$ and $\lit' \OR \beta$ 
where $\lit$ and $\lit'$ are literals for the same variable. What is particularly
striking is this. If the CNF is locally fixated (Def.~\ref{def:fixation}), as is the case for general reasons,
then one can compute the general necessary reasons by simply discarding clauses that are 
not variable-minimal after each resolution step~\cite{corr/abs-2304.14760}. 

\subsection{Targeting a New Decision}
The previous discussion showed how (general) necessary reasons can be used to flip a decision but without a guarantee
on what the new decision is. This becomes an issue when the classifier has more than two classes and hence can
make more than two decisions. Consider a classifier that decides whether an applicant should be approved for 
large loan ($c_1$), approved for a small loan ($c_2$) or declined ($c_3$). Let \(\Delta_1, \Delta_2, \Delta_3\) be
the corresponding class formulas and suppose we have an applicant $\inst$ who was approved for a small loan,
that is, $\inst \models \Delta_2$. The (general) necessary reasons for this decision do suggest minimal changes to the application
that will flip the decision but they do not guarantee whether the new decision will approve a larger loan or decline the application.
Suppose that we wish to flip the decision so the applicant is approved for a larger loan ($c_1$).
What we can do is merge classes \(c_2\) and \(c_3\) leading to a classifier with two decisions \(c_1, c_{23}\)
and corresponding class formulas \(\Delta_1\) and \(\Delta_{23} = \Delta_2 \OR \Delta_3\). The decision
to approve a small loan $(c_2)$ in the original classifier, $\inst \models \Delta_2$, is now a decision to approve a small
loan or decline the application ($c_{23}$) in the new classifier, \(\inst \models \Delta_{23}\). Flipping this
decision is then guaranteed to approve a large loan ($c_1$). 

More generally, suppose we have a classifier with classes \(c_1, \ldots, c_n\), class formulas \(\Delta_1, \ldots, \Delta_n\),
and an instance \(\inst\) in class \(c_i\), \(\inst \models \Delta_i\). If we wish to minimally change this instance so
it moves to some other class \(c_j\), we need to compute the complete reason \(\forall \inst \cdot \sum_{k \neq j} \Delta_k\) or
the general reason \(\iuq \inst \cdot \sum_{k \neq j} \Delta_k\). This can then be followed by computing the (variable-minimal) prime implicates of
these reasons as discussed earlier. This technique of merging class formulas was proposed in~\cite{DarwicheJi22} and the resulting
necessary reasons have been known as \textit{targeted contrastive explanations}~\cite{aiia/IgnatievNA020}.

\section{Compiling Class Formulas}
\label{sec:compilation}

A comprehensive discussion of compiling class formulas for various types of classifiers
is beyond the scope of this tutorial.\footnote{The term ``class formulas'' was first used in~\cite{DarwicheJi22}. 
Compiling the ``decision function'' or ``input-output behavior'' of a classifier are more common terms.} 
However, we will share some key insights about this
compilation process to make it somewhat less mysterious.
The first observation is that the techniques underlying this process depend on the type of classifier,
and its difficulty depends on both the type of classifier and the desired form of compiled formulas.
A number of works have targeted compilations in the form of \textit{tractable circuits}~\cite{pods/Darwiche20} but we
will largely ignore this dimension here since compiling formulas into tractable circuits is a well understood 
problem that has been studied extensively in the area of \textit{knowledge compilation}~\cite{DBLP:journals/jair/DarwicheM02}. 

\subsection{Bayesian Networks}

\begin{figure}[tb]
\centering
\includegraphics[width=9.0cm]{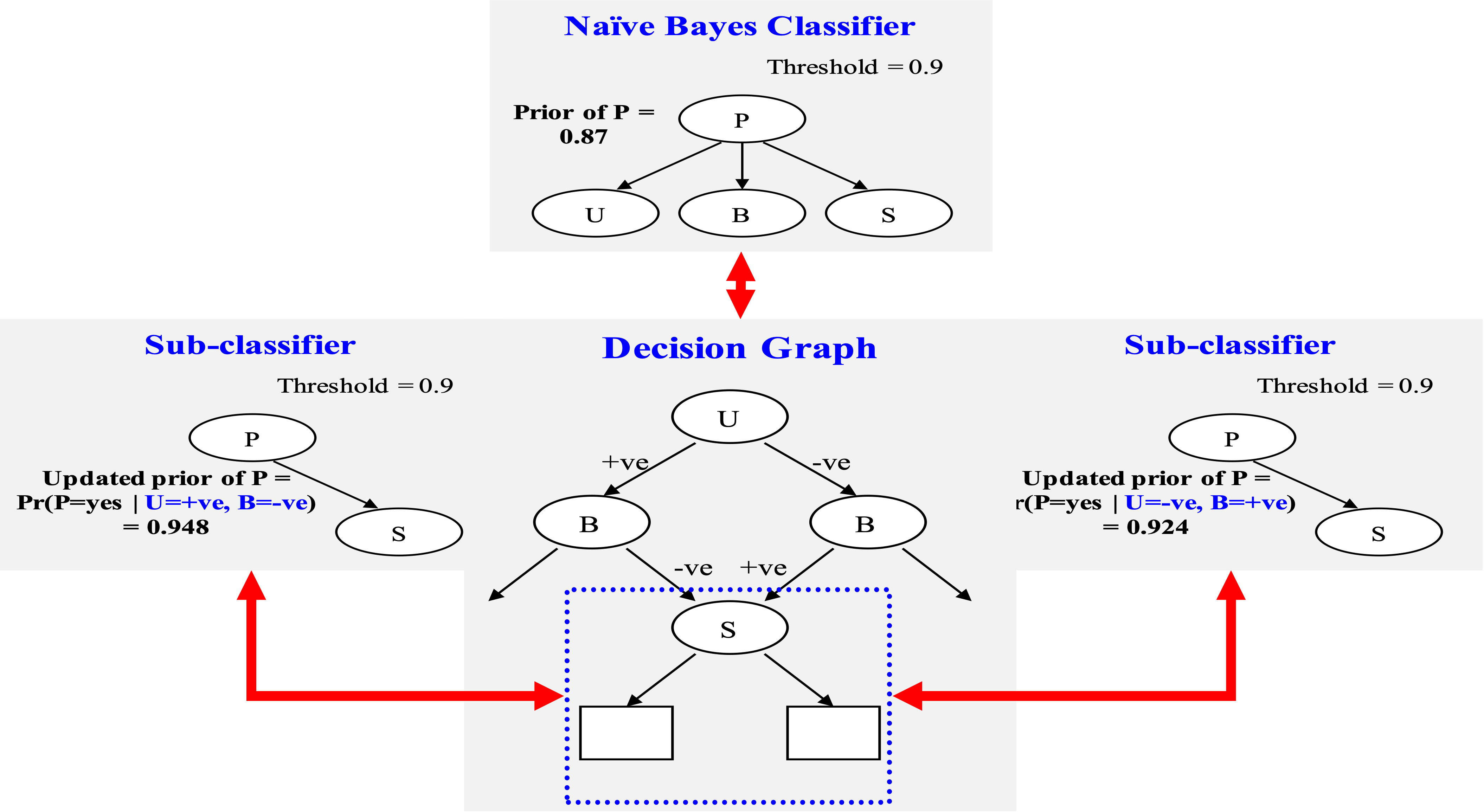}
\caption{Compiling a Na\"ive Bayes classifier into a decision graph~\cite{uai/ChanD03}.}
\label{fig:compile-nbc}
\end{figure}

Compiling the class formulas of Bayesian network classifiers is perhaps the most subtle conceptually since deciding
what instance belongs to what class requires probabilistic reasoning. 
There are two fundamental insights behind this compilation process. The first is the notion of \textit{equivalence intervals} introduced 
in~\cite{uai/ChanD03}. Consider the Na\"ive Bayes classifier in Fig.~\ref{fig:nbc} which has the class
distribution $\Pr(\eql{P}{\yes}),\Pr(\eql{P}{\no}) = (0.87,0.13)$. We can change this distribution quite significantly
without changing any decision made by this classifier as long as $\Pr(\eql{P}{\yes}) \in [0.684,0970]$, which is
called an equivalence interval. The second fundamental insight is the notion of \textit{equivalent sub-classifiers}
also introduced in~\cite{uai/ChanD03}. Continuing with the same classifier, suppose we set the features
$U, B$ to values $\Pos, \Neg$. This leads to a sub-classifier over one
features $S$ with a new class distribution $(0.948,0.052)$ since 
$\Pr(\eql{P}{\yes}|\eql{U}{\Pos},\eql{B}{\Neg}) = 0.948$. Similarly, if these two features are set to $\Neg, \Pos$, we get
another sub-classifier over the same features $U$ but with a different class distribution $(0.924,0.076)$ 
since $\Pr(\eql{P}{\yes}|\eql{U}{\Neg},\eql{B}{\Pos}) = 0.924$.
By utilizing the concept of equivalence intervals, we can conclude
that these two sub-classifiers are equivalent, that is, they make the same decisions on sub-instances $\eql{S}{\Pos}$ and $\eql{S}{\Neg}$.
Putting these two insights together, we can compile a Bayesian network classifier into a (symbolic) decision 
graph by conducting a depth-first search on the space of feature instantiations while pruning the search whenever
we encounter a sub-classifier that is equivalent to another sub-classifier that we already encountered (compiled); see Fig.~\ref{fig:compile-nbc}.
This was used to compile Na\"ive Bayes classifiers in~\cite{uai/ChanD03}, tree-structured classifiers in~\cite{ijcai/ShihCD18} and
graph-structured classifiers in~\cite{aaai/ShihCD19}. The
techniques for computing equivalence intervals and for identifying equivalent sub-classifiers depend on the classifier's structure which explains this progression. 
Fig.~\ref{fig:nbc-dgraph} depicts the compiled decision graph using this method for the Na\"ive Bayes classifier in Fig.~\ref{fig:nbc}.
Extracting class formulas from decision graphs is discussed next.

\subsection{Decision Graphs and Random Forests}

Our next discussion on decision graphs applies to decision trees since they are a special case.
It is well known that the class formulas of decision graphs can be directly obtained as DNFs. 
To construct the DNF for a class \(c_i\), we simply construct a term for each path from the root to a leaf labeled with \(c_i\)
and then disjoin these terms. 
For example, in the decision graph below, there are two paths from the root to class \(c_1\)  which generate the DNF
\(\Delta_1 = x_{12}  \OR x_3 \AND y_1 \AND z_{13}\).
\begin{wrapfigure}[9]{r}{0.18\textwidth}
\centering
 \scalebox{0.75}{
        \begin{tikzpicture}[
        roundnode/.style={circle ,draw=black, thick},
        squarednode/.style={rectangle, draw=black, thick},
        ]
        \node[squarednode]     (X)                              {\Large X};
        \node[roundnode]     (c1)       [below=of X, xshift = -0.8cm, yshift = 0.5cm] {\Large $c_1$};
        \node[squarednode]     (Y)       [below=of X, xshift = 0.8cm, yshift = 0.5cm] {\Large $Y$};
        \node[squarednode]       (Z1)   [below=of Y, xshift = -0.8cm, yshift = 0.5cm] {\Large $Z$};
        \node[squarednode]       (Z2)   [below=of Y, xshift = 0.8cm, yshift = 0.5cm] {\Large $Z$};
        \node[roundnode]       (c2)   [below=of Z1, xshift = -0.8cm, yshift = 0.5cm] {\Large $c_1$};
        \node[roundnode]       (c3)   [below=of Z1, xshift = 0.8cm, yshift = 0.5cm] {\Large $c_2$};
        \node[roundnode]       (c4)   [below=of Z2, xshift = 0.8cm, yshift = 0.5cm] {\Large $c_3$};
        
        \draw[-latex, thick] (X.240) -- node [anchor = center, xshift = -4mm, yshift = 1mm] {$x_1x_2$} (c1.north);
        \draw[-latex, thick] (X.300) -- node [anchor = center, xshift = 4mm, yshift = 1mm] {$x_3$} (Y.north);
        \draw[-latex, thick] (Y.240) -- node [anchor = center, xshift = -4mm, yshift = 1mm] {$y_1$} (Z1.north);
        \draw[-latex, thick] (Y.300) -- node [anchor = center, xshift = 4mm, yshift = 1mm] {$y_2y_3$} (Z2.north);
        \draw[-latex, thick] (Z1.240) -- node [anchor = center, xshift = -4mm, yshift = 1mm] {$z_1z_3$} (c2.north);
        \draw[-latex, thick] (Z1.300) -- node [anchor = center, xshift = -4mm] {$z_2$} (c3.north);
        \draw[-latex, thick] (Z2.240) -- node [anchor = center, xshift = 4mm] {$z_2$} (c3.north);
        \draw[-latex, thick] (Z2.300) -- node [anchor = center, xshift = 4mm, yshift = 1mm] {$z_1z_3$} (c4.north);
        \end{tikzpicture}
        }
\end{wrapfigure}
For decision graphs (that are not trees), this may yield a DNF that is exponentially larger than the graph. 
We can alleviate this by constructing an NNF circuit whose size is guaranteed to be linear in the decision
graph size. The next construction from~\cite{DarwicheJi22} produces circuits that satisfy even stronger
properties as such circuits will allow one, under weak conditions, 
to compute complete and general reasons in time linear in the circuit size.

Let \(c_1, \ldots, c_n\) be the classes of the decision graph (i.e., labels of leaf nodes) and suppose we wish
to construct an NNF circuit for the formula of some class \(c_k\). We first define the function $\texttt{nnf}(N)$ which maps
a node $N$ in the decision graph into an NNF fragment as follows. 
If node $N$ has outgoing edges $\xrightarrow{\lit_1} C_1, \ldots, \xrightarrow{\lit_m} C_m$,
then $\texttt{nnf}(N) = \BOR_{i=1}^m  (\lit_i \AND \texttt{nnf}(C_i))$. 
For leaf nodes, $\texttt{nnf}(c_i) = \top$ if $c_i = c_k$ and $\texttt{nnf}(c_i) = \bot$ if $c_i \neq c_k$.
We can now convert the decision graph into an NNF circuit by calling $\texttt{nnf}(R)$
where $R$ is the graph's root node. This is the standard method but~\cite{DarwicheJi22} defined 
the function $\texttt{nnf}(.)$ differently so the NNF circuits satisfy desirable properties. 
In particular, for an internal node $N$,
it instead used $\texttt{nnf}(N) = \BAND_{i=1}^m  (\lit'_i \OR \texttt{nnf}(C_i))$ where literal \(\lit'_i\) is the complement of literal \(\lit_i\). 
Leaf nodes are kept the same, $\texttt{nnf}(c_i) = \top$ if $c_i = c_k$ and $\texttt{nnf}(c_i) = \bot$ if $c_i \neq c_k$.
The new method is equivalent to using the first method to construct an NNF circuit for the union of
classes $c_1, \ldots, c_{k-1}, c_{k+1}, \ldots, c_n$ and then negating the resulting circuit using deMorgan's law.

The NNF circuits constructed by this new method are guaranteed to be or-decomposable (Def.~\ref{def:or-decomposable}) 
if the decision graph satisfies the \textit{test-once property} (a feature is tested at most once on any path).
Recall that complete and general reasons can be computed in linear time if the class formulas are or-decomposable NNFs
since the  operators $\forall$~(Def.~\ref{def:forall}) and $\iuq$~(Def.~\ref{def:iuq}) will distribute over disjunctions and conjunctions in such NNFs. 
One can also obtain NNF circuits that allow linear-time
computation of complete and general reasons if the decision graph satisfies the \textit{weak test-once property}
discussed in~\cite{DarwicheJi22}. Discretized decision graphs satisfy this property (e.g., the decision tree in Fig.~\ref{fig:dtree-d}).
The above construction and its variants are the basis for the closed-form complete reasons proposed
in~\cite{DarwicheJi22} and the closed-form general reasons in~\cite{corr/abs-2304.14760}.

For random forests with majority voting, one can easily construct NNF circuits for class formulas by combining the NNF circuits for trees
in the forest using a majority circuit. But the resulting circuit is not guaranteed to be or-decomposable even when the circuits for trees are or-decomposable.

\subsection{Neural Networks}

The compilation of neural network classifiers into class formulas is more involved due to the multiplicity of techniques
and assumptions such as the type of activation functions and whether the network is binary, binarized or quantized. We will therefore
restrict the discussion to one approach for binary neural networks while giving pointers to other approaches. 

The work we shall sample is~\cite{kr/ShiSDC20} which assumed neural networks with binary inputs and step-activation functions.
The compilation technique is based on a few observations. First, a neuron with step activation has a binary output so if
its inputs are also binary then the neuron represents a Boolean function.
Hence, a neural network with binary inputs and step-activation functions must represent a Boolean function (i.e., the
signals on its inputs, internal wires and output must all be in $\{0,1\}$). One can therefore convert the neural network
into a Boolean circuit (from which class formulas can be easily extracted) if one can compile a neuron into a Boolean
circuit (i.e.,~one with binary inputs and a step-activation function). The next observation is that a neuron with step activation
is a linear classifier similar to Na\"ive Bayes classifiers. Hence, the technique we discussed
earlier for compiling Na\"ive Bayes classifiers into decision graphs, from~\cite{uai/ChanD03}, can be adopted for compiling 
this class of neurons into decision graphs and then NNF circuits. Once such neurons are successfully compiled,
the neural network can be immediately represented as a Boolean circuit from which class formulas can be easily obtained.
This work went a bit further by employing a variant on the compilation method in~\cite{uai/ChanD03} which assumes that
the neuron weights \(w_1, \ldots, w_n\) and threshold \(T\) are integers. This allows one to conduct the compilation
in $O(nW)$ time where \(W = |T|+\sum_{i=1}^n |w_i|\) is a sum of absolute values. This pseudo-polynomial time
compilation algorithm can be applied to real-valued weights by multiplying the weights by a constant and
then truncating (i.e., the weights have fixed precision). This technique permitted the compilation of neural networks with hundreds of
inputs (features). The example in Fig.~\ref{fig:digits} and many other examples in~\cite{kr/ShiSDC20} were produced by this approach.

For further techniques that encode input-output behavior symbolically, 
see~\cite{aaai/NarodytskaKRSW18,iclr/NarodytskaZGW20,sat/NarodytskaSMIM19,ccs/BalutaSSMS19,sat/ShihDC19,nips/KaiM20,cav/ZhangZCSC21} 
for binarized neural networks 
and~\cite{tacas/GiacobbeHL20,aaai/HenzingerLZ21,acm/ZhangZCSZCS23,corr/abs-2212-02781} for quantized ones.
Most of these works target verification tasks though instead of explaining behavior.

\section{Concluding Remarks}
\label{sec:conclusion}

We close this tutorial with three remarks. 
The first remark concerns two distinct classes of works on explainable AI. 
One class assumes enough information about the classifier
to allow the construction of class formulas or, more generally, the characterization of which instances belong to what classes.
This is clearly the assumption we made in this tutorial and the resulting approaches are known as \textit{model-based.}
These approaches usually seek explanations that come with hard guarantees like the ones we discussed.
The other class of works assume that we can only query the classifier, that is, ask it to classify instances. These approaches
are known as \textit{model-agnostic} and have been mostly popularized through the early systems described 
in~\cite{LIME,ANCHOR}. These approaches tend to scale better but do not offer hard guarantees and can be viewed
as computing approximate explanations~\cite{JoaoApp}.

The second remark relates to the extensive nature of investigations that were conducted on explainability over the last few years, 
particularly on the complexity of computing explanations. The following are some examples.
For Na\"ive Bayes (and linear) classifiers, it was shown that one sufficient reason can be generated in log-linear time, 
and all sufficient reasons can be generated with polynomial delay~\cite{nips/0001GCIN20}. 
For decision trees, the complexity of generating one sufficient reason was shown to be in polynomial time~\cite{corr/abs-2010-11034}.
Later works showed the same complexity for decision graphs~\cite{DBLP:conf/kr/HuangII021} and some classes
of tractable circuits~\cite{kr/AudemardKM20,corr/abs-2107-01654}.
The generation of sufficient reasons for decision trees was also studied in~\cite{Audemard2021OnTE}, 
including the generation of shortest sufficient reasons which was shown to be hard even for a single reason.
The generation of shortest sufficient reasons was also studied in a broader
context that includes decision graphs and SDDs~\cite{DarwicheJi22}.\footnote{SDDs ({\em Sentential Decision Diagrams}) are decision diagrams that branch
on formulas (sentences) instead of variables~\cite{ijcai/Darwiche11}.
SDDs are a superset of, and exponentially more succinct than~\cite{aaai/Bova16}, OBDDs but are not comparable to some other types of decision graphs
in terms of succinctness~\cite{mst/BolligB19,uai/BeameL15,uai/BeameLRS13}.}
The complexity of shortest sufficient reasons was studied in~\cite{nips/BarceloM0S20}
for Boolean classifiers which correspond to decision graphs and for neural networks with ReLU activation functions. 
It was further shown that 
the number of necessary reasons is linear in the decision tree size~\cite{DBLP:conf/kr/HuangII021,DarwicheJi22}, that
all such reasons can be computed in polynomial time~\cite{Audemard2021OnTE,DarwicheJi22}, and that
the shortest necessary reasons can be enumerated with polynomial delay if the classifier satisfies some conditions as stated in~\cite{KR2020-86}.
Further complexity results were shown in~\cite{kr/AudemardKM20,corr/abs-2107-01654},
where classifiers where categorized based on the tractable circuits that represent them~\cite{corr/abs-2107-01654}
or the kinds of processing they permit in polynomial time~\cite{kr/AudemardKM20}.
A comprehensive study of complexity was also presented in~\cite{DBLP:conf/kr/AudemardBBKLM21} for a large set of explanation
queries and classes of Boolean classifiers.
Computational approaches based on either SAT, MaxSAT or partial MaxSAT were also proposed for
random forests, e.g.,~\cite{DBLP:conf/ijcai/Izza021,DBLP:conf/aaai/AudemardBBKLM22}, 
tree ensembles, e.g.,~\cite{DBLP:conf/aaai/IgnatievIS022} and
boosted trees, e.g.,~\cite{DBLP:journals/corr/abs-2209-07740}.
Formal results on explainability were even employed to question common wisdoms like those relating to the interpretability
of decision trees~\cite{DBLP:journals/jair/IzzaIM22}.

The last remark relates to the storyline adopted in this tutorial which treated the  complete and general reasons behind decisions
as the core notions in this theory of explainability and used them to describe other notions, even ones that
were proposed before such reasons were conceived. 
This is an outcome of our firm belief that the notion of ``instance abstraction'' 
must be the core of any comprehensive and well-founded theory of explainability.
We hope the reader would agree with us that this treatment has also led to a minimalistic formulation that explicates
semantics and facilitates computation.

\section*{Acknowledgment}
I wish to thank Yizuo Chen, Haiying Huang and Albert Ji for their useful feedback and discussions.

\bibliographystyle{IEEEtran}
\bibliography{LICS,bibliography1,bibliography2}

\end{document}